\documentclass[letterpaper, 10 pt, journal, twoside]{IEEEtran}
\usepackage{mathptmx} 
\usepackage{amsmath} 
\usepackage{amssymb}  
\usepackage{bm}
\usepackage{esvect}
\usepackage{comment}
\usepackage{graphicx}
\usepackage{cite}
\graphicspath{./Images}
\graphicspath{./Layout}

\usepackage{xcolor}

\usepackage{algorithm}
\usepackage{algorithmic}
\usepackage{url}

\begin{document}

\title{A Self-Tuning Impedance-based Interaction Planner \\for Robotic Haptic Exploration
}

\author{Yasuhiro Kato$^{1}$, Pietro Balatti$^{2}$~\IEEEmembership{Member, IEEE}, Juan M. Gandarias$^{2}$~\IEEEmembership{Member, IEEE}, Mattia Leonori$^{2}$,\\ Toshiaki Tsuji$^{1}$~\IEEEmembership{Member, IEEE} and Arash Ajoudani$^{2}$~\IEEEmembership{Member, IEEE}
\thanks{Manuscript received: February, 24, 2022; Revised April, 28, 2022; Accepted June, 27, 2022.}
\thanks{This paper was recommended for publication by Editor Clement Gosselin upon evaluation of the Associate Editor and Reviewers' comments.} 
\thanks{This work was supported in part by the European Union’s Horizon 2020 research and innovation programme under Grant Agreement No. 871237 (SOPHIA) in part by the ERC-StG Ergo-Lean (Grant Agreement No.850932).}
\thanks{$^{1}$Y. Kato (corresponding author) and T. Tsuji are with Graduate School of Science and Engineering,
        Saitama University, 255 Shimo-Okubo, Sakura-ku, Saitama, Japan
        {\tt\small \{y.kato.421, tsuji\}@mail.saitama-u.ac.jp}}%
\thanks{$^{2}$P. Balatti, J. M. Gandarias, M. Leonori and A. Ajoudani are with HRI$^{2}$ Lab of the Istituto Italiano di Tecnologia, 16163 Genoa, Italy
        {\tt\small \{pietro.balatti, juan.gandarias, mattia.leonori, arash.ajoudani\}@iit.it}}%
\thanks{Digital Object Identifier (DOI): see top of this page.}
}

\markboth{IEEE Robotics and Automation Letters. Preprint Version. Accepted June, 2022}
{Kato \MakeLowercase{\textit{et al.}}: A Self-Tuning Impedance-based Interaction Planner}


\maketitle

\begin{abstract}
This paper presents a novel interaction planning method that exploits impedance tuning techniques in response to environmental uncertainties and unpredictable conditions using haptic information only. The proposed algorithm plans the robot’s trajectory based on the haptic interaction with the environment and adapts planning strategies as needed. Two approaches are considered: \textit{Exploration} and \textit{Bouncing} strategies. The \textit{Exploration} strategy takes the actual motion of the robot into account in planning, while the \textit{Bouncing} strategy exploits the forces and the motion vector of the robot. Moreover, self-tuning impedance is performed according to the planned trajectory to ensure compliant contact and low contact forces. In order to show the performance of the proposed methodology, two experiments with a torque-controller robotic arm are carried out. The first considers a maze exploration without obstacles, whereas the second includes obstacles. The proposed method performance is analyzed and compared against previously proposed solutions in both cases. Experimental results demonstrate that: i) the robot can successfully plan its trajectory autonomously in the most feasible direction according to the interaction with the environment, and ii) a compliant interaction with an unknown environment despite the uncertainties is achieved. Finally, a scalability demonstration is carried out to show the potential of the proposed method under multiple scenarios.
\end{abstract}



\section{Introduction}
\IEEEPARstart{F}{lexible} materials naturally deform when interacting with the surrounding environment. This behavior enables us to exploit environmental constraints, facilitating `blind' explorations through physical interaction, e.g., insertion tasks~\cite{klein2001robot, chentanez2009interactive} or pipe inspection~\cite{nassiraei2007concept}. For instance, running a wire inside the cable raceways hidden in a wall becomes a very simple task, which can be accomplished by using a flexible stick and pushing it from an external tip. In other words, the elastic behavior allows solving the maze even without `planning' the trajectory. Such elastic behavior (i.e., reacting and accommodating to changes in the environment) can be replicated by a robot in autonomous tasks despite existing uncertainties and missing of visual data.

Traditional solutions rely on trajectory planners to deal with uncertainties, avoiding unexpected contacts with obstacles such as RRT~\cite{kuffner2000rrt} or PRM~\cite{kavraki1996probabilistic}. These approaches need to build or estimate a highly accurate environment model. However, the more complex the environment, the more challenging is the modeling of the environment, which makes these solutions impractical. 
On the other hand, allowing contacts with the environment and planning based on haptic information enables the robot to perform better and more autonomously in complex environments. In this context, planners based on contact forces have been proposed~\cite{pall2018contingent, yang2018dmps}. Although the environmental uncertainty is significantly reduced by considering contact, those planning methods are computationally expensive and require building/having a map of the environment. Trajectory-scaling~~\cite{haddadin2008collision} or admittance behavior~\cite{de2006collision} are computationally less expensive and can flexibly react to the external world. However, these solutions may not be suitable for exploration/path-finding tasks in which the environment can change significantly.

\begin{figure}[t]
  \centering
  \includegraphics[width=0.84\columnwidth]{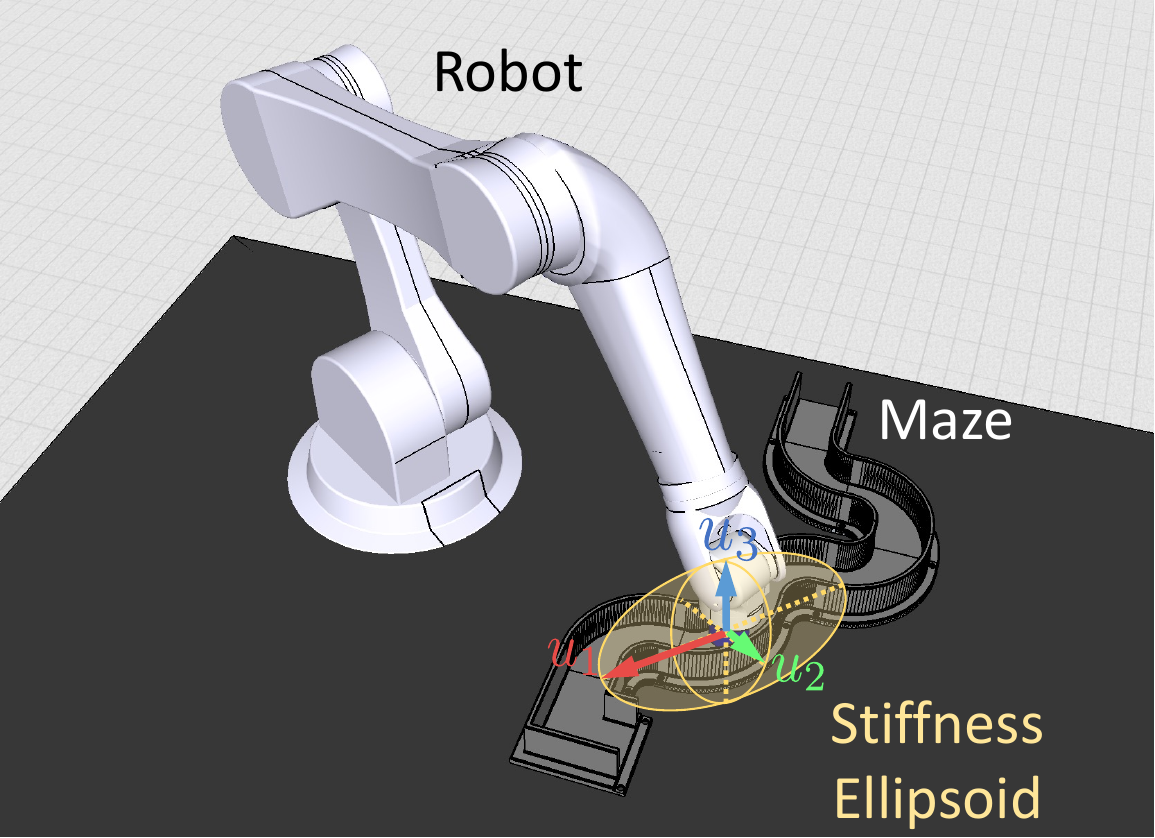}
  \caption{Maze exploration concept: the robotic arm carries out an exploration task in an unknown and constrained environment, planning the trajectories while interacting with it. The impedance parameters are adaptively tuned online according to the direction of the motion: the robot is stiffer along the principal axis of the stiffness ellipsoid ($u_1$), and more compliant in the others ($u_2, u_3$). The arrows' length represent the Cartesian stiffness and damping values.}
  \label{fig:intro}
  \vspace{-7mm}
\end{figure}

Regarding the problem of physical interaction with unstructured environments, impedance controllers~\cite{hogan1985impedance} are known to be an effective way to regulate the interaction forces when a contact occurs. In particular, variable impedance control~\cite{tsumugiwa2002variable,ikeura1995variable} can yield soft contacts between the robot and the environment/human when necessary. This can be advantageous in human-collaborative scenarios, since the compliance introduced by the controller allows the person to modify the robot's trajectory~\cite{nemec2018human}. On the other hand, in the impedance control scheme, high stiffness may be required to achieve high precision or interaction forces~\cite{ajoudani2012tele, abu2018force, ajoudani2018reduced}. For instance, in a collaborative sawing task~\cite{peternel2016towards}, the stiffness values were higher in the direction of sawing, to create cutting forces, and lower in others, to comply with the task constraints. Therefore, it is necessary to tune the impedance profiles of the robots based on the actual task, human, and environmental states. In such a way, robots can respond to modeling/environment uncertainties or unexpected changes~\cite{mobedi2020adaptive,gribovskaya2011motion}.

This paper proposes a novel interaction planner that shapes the trajectory and the direction of maximum stiffness without prior knowledge of the environment, only exploiting real-time haptic interaction, being naturally compliant to the environmental constraints. This method regulates the haptic interaction with the external world by an active impedance control scheme. One of the most significant benefits of our method is that it does not require prior knowledge/map of the environment. Furthermore, our interaction planner adapts two planning strategies based on the robot state. The \textit{Exploration} strategy plans by considering the actual concurrent motion of the robot, while the \textit{Bouncing} strategy employs the interaction forces and previous motion vector of the robot.
On the other hand, the second strategy reacts when the robot gets stuck by the environmental constraints. The main components of the proposed approach are the novel interaction planner and our self-tuning impedance technique, previously presented in~\cite{balatti2020method}. The interaction planner, which deploys the robot's motion and estimated forces, is responsible for planning the exploration trajectory by interacting with the environment. The self-tuning impedance method renders a stiffer profile in the direction of its primary motion while still being compliant along the other axes. This method allows the robot to adapt to environmental changes and avoid unnecessarily high interaction forces when in contact with rigid materials.

We validate the proposed approach with experiments on a real system using the Franka Emika Panda robotic arm. To do so, we built a modular maze to create different paths with rigid environmental constraints to evaluate the robot's performance during autonomous path generation and tracking. The proposed method was additionally compared to fully stiff and compliant cases using the same \textit{Exploration} and \textit{Bouncing} logics.

The paper is structured as follows: Sec.~\ref{sec:preliminary} and \ref{sec:method} present preliminaries and the proposed interaction planning strategy. Sec.~\ref{sec:experiments} discusses the experimental setup and the experiments. Finally, discussion and conclusion are addressed in Sec.~\ref{sec:conclusion}.

\section{Preliminary}\label{sec:preliminary}

\subsection{Cartesian impedance controller}
An active impedance control can be implemented through torque sensing and actuation~\cite{hogan1987stable}, with the robot joint torques $\bm{\tau}\in\mathbb{R}^{n}$ calculated as:
\begin{equation}
    \bm{\tau} = \bm{M(q)\ddot{q}}+\bm{C(q,\dot{q})\dot{q}}+\bm{g(q)}+\bm{\tau}_{\textrm{ext}},
\end{equation}

\begin{equation}
    \bm{\tau}_{\textrm{ext}} = \bm{J(q)}^{T}\bm{F}_{c}+\bm{\tau}_{\textrm{st}},
\end{equation}
where n is the number of joints, $\bm{q}\in\mathbb{R}^{n}$ is the joint angles vector, $\bm{J}\in\mathbb{R}^{6 \times n}$ is the Jacobian matrix, $\bm{M}\in\mathbb{R}^{n \times n}$ is the mass matrix, $\bm{C}\in\mathbb{R}^{n \times n}$ is the Coriolis and centrifugal matrix, and $\bm{g}\in\mathbb{R}^{n}$ is the gravitational vector. $\bm{\tau}_{\textrm{ext}}\in\mathbb{R}^{n}$ represents the external torques and $\bm{\tau}_{\textrm{st}}\in\mathbb{R}^{n}$ is the secondary task torques in the null-space of $\bm{J}$. The Cartesian forces $\bm{F}_c\in\mathbb{R}^{6}$ are calculated as:
\begin{equation}
    \bm{F}_c = \bm{K}_{c}(\bm{x}_{d}-\bm{x}) + \bm{D}_{c}(\bm{\dot{x}}_{d}-\bm{\dot{x}}),
\end{equation}
where $\bm{K}_c\in\mathbb{R}^{6 \times 6}$ and $\bm{D}_{c}\in\mathbb{R}^{6 \times 6}$ represent, respectively, the Cartesian stiffness and damping matrix. $\bm{x}_d$ and $\bm{x}\in\mathbb{R}^{6}$ are the Cartesian desired and actual poses, while $\bm{\dot{x}}_d$ and $\bm{\dot{x}}\in\mathbb{R}^{6}$ represent their corresponding velocity profiles.

\subsection{Self-tuning impedance unit}
Hereafter, we introduce the basic principles of the self-tuning impedance controller, originally presented in~\cite{balatti2020method}, that is going to be employed with the method proposed in this work. This adaptive controller allows the robot to accurately track the desired motion along the motion vector, but also allows flexibility along the other directions to make the robot adapt to external unintended disturbances (e.g., obstacles). Being more stiff in the principal direction of the movement, and more compliant in the others can be achieved by tuning the major axis of the Cartesian stiffness and damping ellipsoids in the direction of interaction, i.e., deriving the Cartesian stiffness and damping matrices (symmetric and positive definite) as follows:
\begin{equation}\label{eq9}
    \bm{K}_{c} = \bm{U}\bm{\Sigma}_{k}\bm{U}^{T},
\end{equation}
\begin{equation}\label{eq10}
    \bm{D}_{c} = \bm{U}\bm{\Sigma}_{d}\bm{U}^{T},
\end{equation}
where the diagonal matrix $\bm{\Sigma}_{k}$ and $\bm{\Sigma}_{d}$ are the desired stiffness and damping factors along the direction of the vector composing the orthonormal basis $\bm{U}$, whose first column represents the motion vector, calculated as $ \boldsymbol{x_{d,t}} - \boldsymbol{x_{d,t-1}}$ and defined as $u_1$ in Fig.~\ref{fig:intro}, and the remaining ones (i.e., $u_2$ and $u_3$) are shaped so as to compose an orthonormal basis. $\bm{\Sigma}_{k}$ and $\bm{\Sigma}_{d}$ are defined by:
\begin{equation}\label{eq11}
    \bm{\Sigma}_{k} = \textrm{diag}(k_{\textrm{max}},k_{\textrm{min}},k_{\textrm{min}}),
\end{equation}
\begin{equation}\label{eq12}
    \bm{\Sigma}_{d} = \textrm{diag}(d_{\textrm{max}},d_{\textrm{min}},d_{\textrm{min}}),
\end{equation}
where $k_{\textrm{max}}$ is the maximum controllable stiffness and $d_{\textrm{max}} = 2\zeta\sqrt{k_{\textrm{max}}}$ \cite{albu2003cartesian} its corresponding damping value. Likewise, $k_{\textrm{min}}$ and $d_{\textrm{min}}$ are the minimum stiffness and damping coefficients.

\section{Method}\label{sec:method}

\begin{figure}
  \centering
  \includegraphics[trim=0.0cm 0.3cm 0.0cm 0.0cm,clip,width=0.85\columnwidth]{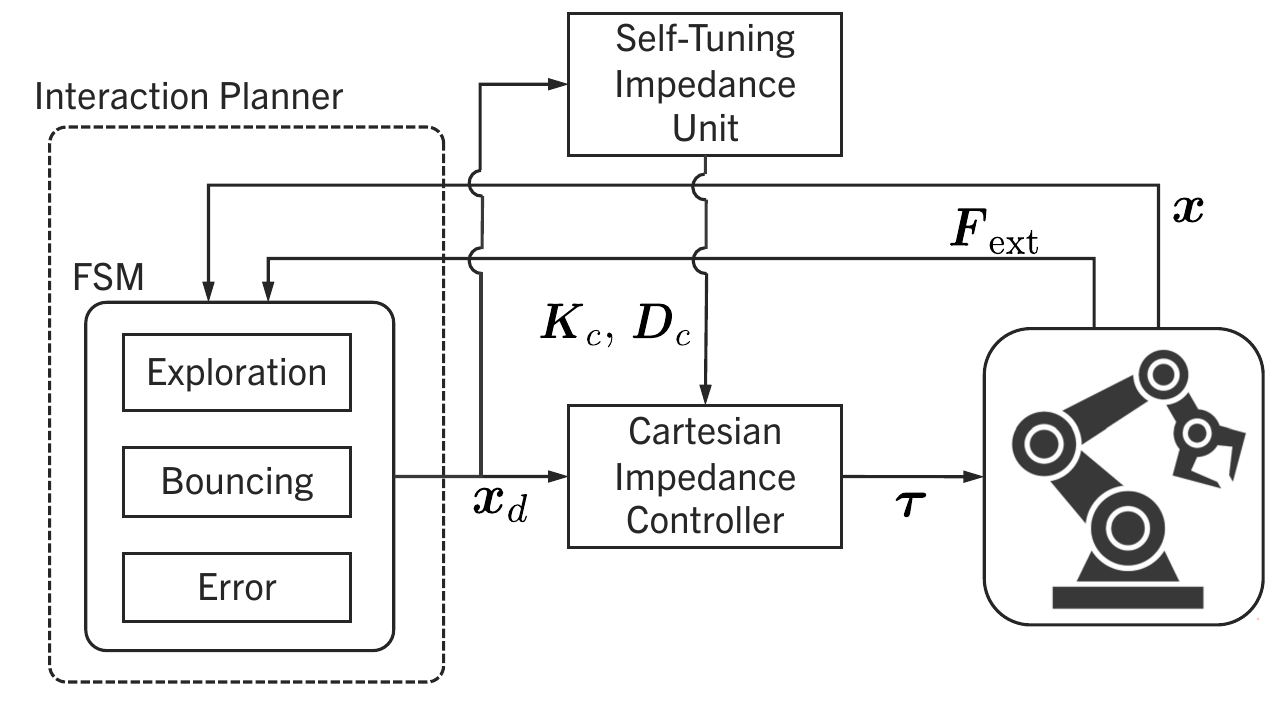}
  \caption{System framework: The framework is composed of 3 modules, Cartesian impedance controller, self-tuning impedance unit, and interaction planner. The self-tuning impedance unit and interaction planner rely on sensing robot state, i.e., external forces and robot pose.}
  \label{SystemConfig}
  \vspace{-6mm}
\end{figure}

\subsection{Interaction planner}
In this study, we consider the case where the robot is not given any prior knowledge about the environment it needs to interact with. Additionally, no external sensors are used (e.g., camera or depth sensor), and the only information the system can rely on are represented by the robot's internal torque sensing, which is used to estimate the external interaction forces. Considering these conditions, in this work we propose an interaction planner that is able to plan the desired trajectories based on the interaction with the environment. The proposed interaction planner mainly consists of two strategies, i.e., \textit{Exploration} and \textit{Bouncing}. The robot autonomously switches between the two, based on the sensed contact forces.
This planner was inspired by~\cite{haddadin2010real}, where planning trajectories are achieved by scaling the desired velocity based on contact forces. Such approach enables flexible trajectory planning without prior knowledge on the external environments. We have exploited and extended this algorithm for the presented \textit{Bouncing} strategy. This strategy has a wider range of scaling factors $[-1...1]$, which allows us not only to avoid obstacles, but also to plan a trajectory towards the goal flexibly by considering interaction forces and motion vector.

\subsubsection{Exploration} 
\begin{figure}
  \centering
  \includegraphics[trim=0.0cm 0.0cm 0.0cm 0.0cm,clip,width=0.45\textwidth]{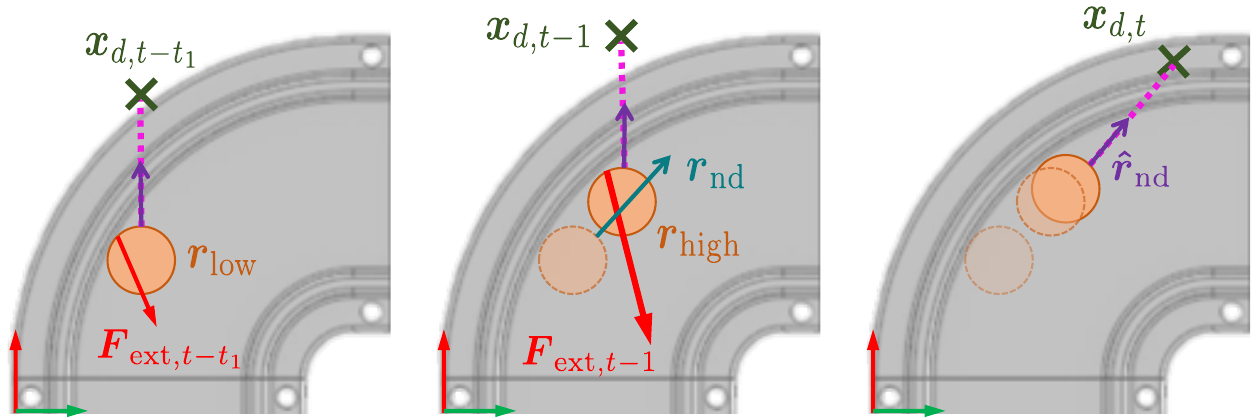}
  \caption{Interaction planner \textit{Exploration} state: (left) shows the phase when $\bm{F}_{\textrm{ext}}$ exceeds $F_{\textrm{th\_low}}$, and thus $\bm{r}_{\textrm{low}}$ is stored. (middle) shows the phase when $\bm{F}_{\textrm{ext}}$ exceeds $F_{\textrm{th\_high}}$, and thus where $\bm{r}_{\textrm{high}}$ is stored. (right) shows the new motion, $\bm{\hat{r}}_{\textrm{nd}}$, based on the displacement caused by the contacts with the environment.}
  \label{fig:exploration}
  \vspace{-2mm}
\end{figure}

in this state, the robot is in free space or, if in contact with the environment, is still able to move along the desired motion (subject to the relative environmental perturbations). The desired direction of the motion is modified only when the contact with the environment reaches the values of certain force thresholds, meaning that the robot needs to deviate its trajectory to be able to explore the surrounding environment, as illustrated in Fig.~\ref{fig:exploration}.

To this end, the \textit{Exploration} algorithm, whose pseudo-code is reported in Alg.~\ref{alg:exploration}, calculates the new desired pose $\bm{x}_d$ based on the actual pose $\bm{x}$ and on the sensed external forces $\bm{F}_{\textrm{ext}}$. The algorithm is in charge of detecting two poses, namely $\bm{r}_{\textrm{low}}$ and $\bm{r}_{\textrm{high}}$, that are then used to compute the new motion direction, $\bm{\hat{r}}_{\textrm{nd}}$ and its relative increment $\bm{\Delta r}_{\textrm{nd}}$. The above-mentioned poses are calculated by assigning the actual robot pose $\bm{x}$ in two different time frames, triggered by two external forces thresholds. The first, $\bm{r}_{\textrm{low}}$, is 
retrieved when the robot external forces go beyond a first threshold named $F_{\textrm{th\_low}}$. While the contact force exceeds $F_{\textrm{th\_low}}$, the algorithm checks if the external forces go beyond the second threshold, $F_{\textrm{th\_high}}$. These thresholds are set empirically. When this is the case, the actual robot pose is assigned to the second pose, $\bm{r}_{\textrm{high}}$. From these values, the new motion vector $\bm{r}_{\textrm{nd}}$ is computed, and then normalized as $\bm{\hat{r}}_{\textrm{nd}}$ so as to avoid significant incremental changes. Lastly, the increment to the previous desired pose, $\bm{x}_{d,t-1}$, is calculated as in line 13, where $v_{\textrm{constE}}$ is the desired velocity, and $\Delta T$ represents control loop time.

\begin{algorithm}[t]
    \caption{Exploration algorithm}\label{alg:exploration}
    \begin{algorithmic}[1]
        \renewcommand{\algorithmicrequire}{\textbf{Input:}}
        \renewcommand{\algorithmicensure}{\textbf{Output:}}
        \REQUIRE ${\bm{x}, \bm{F}_{\textrm{ext}}}$
        \ENSURE  $\bm{x}_{d}$
        
        \textit{Control loop} :
        \IF {($F_{\textrm{th\_high}}$ is not detected)}
            \IF {($F_{\textrm{th\_low}}$ is not detected)}
                \IF {($\bm{F}_{\textrm{ext}} >=  F_{\textrm{th\_low}}$)}
                    \STATE $\bm{r}_{\textrm{low}} =  \bm{x}$ \{{$F_{\textrm{th\_low}}$ is detected}\}
                \ENDIF
            \ENDIF
                \IF {($\bm{F}_{\textrm{ext}} >=  {F}_{\textrm{th\_high}}$)}
                    \STATE $\bm{r}_{\textrm{high}} =  \bm{x}$ \{$F_{\textrm{th\_high}}$ is detected\}
                \ENDIF
        \ENDIF
        \STATE $\bm{r}_{\textrm{nd}} = \bm{r}_{\textrm{high}}-\bm{r}_{\textrm{low}}$
        \STATE $\bm{\hat{r}}_{\textrm{nd}} = \cfrac{\bm{r}_{\textrm{nd}}} {\| \bm{r}_{\textrm{nd}} \|_2} $
        \STATE $\Delta\bm{r}_{d} = \bm{\hat{r}}_{\textrm{nd}} * v_{\textrm{constE}}*\Delta T$
        \STATE $\bm{x}_{d,t} = \bm{x}_{d,t-1} + \Delta\bm{r}_{d}$
    \end{algorithmic}
\end{algorithm}

\subsubsection{Bouncing} 
\begin{figure}[t]
  \centering
  \includegraphics[trim=0.0cm 0.4cm 0.0cm 0.0cm,clip,width=0.45\textwidth]{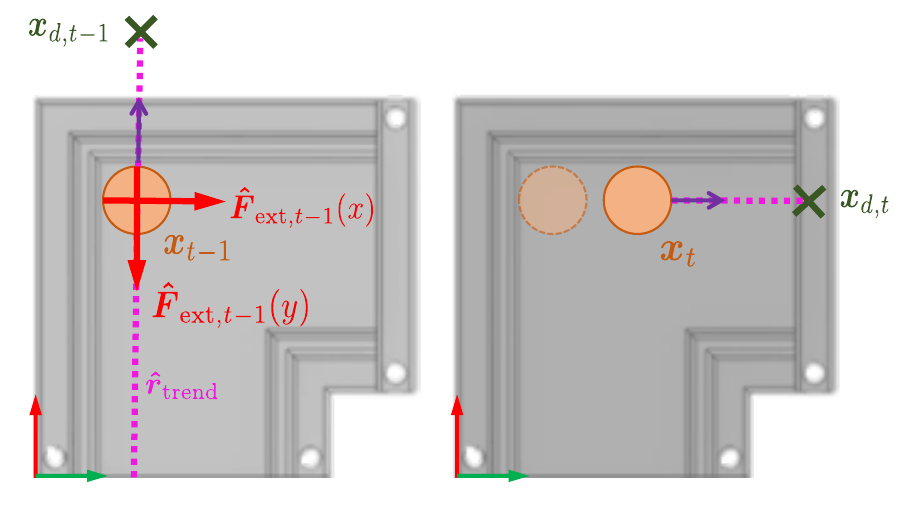}
  \caption{Interaction planner \textit{Bouncing} state: (left) shows the phase where a robot is trapped by environmental constraints. At this point, by considering the sensed external forces and the motion trend vector, the robot finds a new direction as shown in (right).}
  \label{fig:bouncing}
  \vspace{-2mm}
\end{figure}

this second strategy was implemented to substantially change the direction of the motion when the robot gets stuck by environmental constraints. When the robot is in contact with the environment perpendicularly, the \textit{Exploration algorithm} presented above is unable to find the movable direction because it relies on the displacement caused by two different contacts with the environment, in order to plan a new trajectory. As a representative example, we can consider an L-shaped constraint that has limited possible directions of movement, as illustrated in Fig.~\ref{fig:bouncing}.

To address this issue, the new direction that needs to be computed cannot rely anymore on two consecutive poses, but the robot needs to be steered to another direction, i.e., bouncing from the contact point.
To determine the bouncing direction, we consider the normalized external forces vector $\bm{\hat{F}}_{\textrm{ext}}\in\mathbb{R}^{6}$ when the robot gets stuck by environmental constraints, and the unit vector $\hat{\bm{r}}_{\textrm{trend}}\in\mathbb{R}^{3}$ , that is the robot motion vector derived as $\bm{r}_{\textrm{trend}} = \bm{x}_{t} - \bm{x}_{t-\textrm{m}}$, 

where $m$ represents the number of time steps in the past that are considered. The larger this values is set, the bigger the robot's motion trend becomes.
The desired velocity $v_{\textrm{constB}}$, is scaled by the scaling factor $\alpha$, $\beta\in[-1...1]$ which are calculated as line~5 and 9 in Alg.~\ref{alg:bouncing}, where $\phi(x)\in[-\pi...\pi]$ and $\phi(y)\in[-\pi...\pi]$ are angles between $\bm{\hat{F}}_{\textrm{ext}}(x)$ and $\bm{\hat{r}}_{\textrm{trend}}$, as well as $\bm{\hat{F}}_{\textrm{ext}}(y)$ and $\bm{\hat{r}}_{\textrm{trend}}$ respectively. The angle between external forces and motion vector is calculated by:

\begin{equation}\label{eq17}
    \phi = arccos\left(\frac{<\bm{\hat{F}}_{\textrm{ext}},\bm{\hat{r}}_{\textrm{trend}}>}{\|\bm{\hat{F}}_{\textrm{ext}}\|\|\bm{\hat{r}}_{\textrm{trend}}\|}\right).
\end{equation}
Furthermore, the scaling factor was designed to be larger when two vectors are perpendicular while it becomes smaller when two vectors are on the same axis. 

\begin{algorithm}[t]
    \caption{Bouncing algorithm}\label{alg:bouncing}
    \begin{algorithmic}[1]
        \renewcommand{\algorithmicrequire}{\textbf{Input:}}
        \renewcommand{\algorithmicensure}{\textbf{Output:}}
        \REQUIRE $\bm{r}_{\textrm{trend}}, \bm{F}_{\textrm{ext}}$
        \ENSURE  $\bm{x}_{d}$
        
        \textit{Initialization} :
        \STATE $\bm{\hat{F}}_{\textrm{ext}} = \cfrac{\bm{F}_{\textrm{ext}}} {\| \bm{F}_{\textrm{ext}} \|_2} $
        \STATE $\bm{\hat{r}}_{\textrm{trend}} = \cfrac{\bm{r}_{\textrm{trend}}} {\| \bm{r}_{\textrm{trend}} \|_2} $
        \STATE $\phi(x) = getAngleBetween({\bm{\hat{F}}_{\textrm{ext}}(x),\bm{\hat{r}}_{\textrm{trend}}})$
        \STATE $\phi(y) = getAngleBetween({\bm{\hat{F}}_{\textrm{ext}}(y),\bm{\hat{r}}_{\textrm{trend}}})$
        \STATE $\alpha = (1 - |cos(\phi(x))|)$
        \IF{($\bm{\hat{F}}_{\textrm{ext}}(x) < 0.0$)}
            \STATE $\alpha = -\alpha$
        \ENDIF
        \STATE $\beta = (1 - |cos(\phi(y))|)$
        \IF{($\bm{\hat{F}}_{\textrm{ext}}(y) < 0.0$)}
            \STATE $\beta = -\beta$
        \ENDIF
        
        \textit{Control loop} :
        \STATE $\Delta{r_{d}(x)} = \alpha * v_{\textrm{constB}}*\Delta T$
        \STATE $\Delta{{r}_{d}(y)} = \beta * v_{\textrm{constB}}*\Delta T$
        \STATE $\bm{x}_{d,t} = \bm{x}_{d,t-1} + \Delta\bm{r}_{d}$
    \end{algorithmic}
\end{algorithm} 

\subsection{Finite state machine}
A Finite State Machine (FSM) is also implemented to enable the robot to change its strategy according to its state. The FSM consists of the following states: \textit{Exploration}, \textit{Bouncing}, and \textit{Error}. The default robot behavior resides in the \textit{Exploration} state, where the robot can navigate through the environment. On the other hand, the FSM shifts to the \textit{Bouncing} state if the robot is considered to be trapped by the constraints. To switch between the two states, the following equation is introduced: 
\begin{equation}\label{eq:bouncingswitch}
    \Delta{d} = \| \bm{x} - \bm{x}_{t-\textrm{h}} \|_2
\end{equation}
where $\Delta{d}$ is the displacement and $h$ is the sampling time. If $\Delta{d} < R_{\textrm{th}}$, where $R_{\textrm{th}}$ is the displacement threshold, the FSM shifts into the \textit{Bouncing} state to continue the exploration task. After the interaction forces decrease, the FSM switches back to the \textit{Exploration} state once again. Additionally, the \textit{Error} state is implemented since we assume that the interaction between the robot and the rigid environment can generate large interaction forces. When these forces become excessively large, the exploration task gets aborted.

\section{Experiments and Results}\label{sec:experiments}

\subsection{Experiment setup and description}
The software architecture was developed with the robotics middleware Robot Operating System (ROS) using C++ as client library. 
The experimental setup included a Franka Emika Panda robotic arm, three different mazes placed in front of the robot, representing environmental constraints, and a peg (Diameter: 30mm - Length: 55mm) mounted at the robot end-effector. The maze parts and the peg were designed and 3D printed with rigid plastic material (PLA). The mazes were constructed from components that could be disassembled and were built combining linear, curved, and L-shaped parts.

In order to validate the presented method, we conducted several experiments, in which the robot had no prior knowledge about the environment it had to interact with. The first two experiments (Sec.~\ref{subsec:exp_12}) demonstrate the performances of the \textit{interaction planner} with the adaptive impedance controller without and with cluttered conditions, i.e., causing interaction resistance. In the third experiment (Sec.~\ref{subsec:exp_3}), the robot performs the exploration task in other two mazes. The maze designs were different from the one in the first two experiments in order to show the scalability of the proposed method. A video of the experiments is available in the multimedia extension\footnote{The video can also be found at \url{https://youtu.be/DDqosdN2274}}.
    
\subsection{Maze exploration experiment}\label{subsec:exp_12}
In the first experiment, the performance of the maze exploration task was evaluated in three scenarios: high impedance, low impedance and self-tuning impedance conditions. In the first two cases, the diagonal stiffness values were set to constant values on the three Cartesian axes, i.e., $1000$~N/m for the high impedance scenario, and $300$~N/m for the low impedance one. In the latter, the value was selected so as to let the robot follow the desired pose of the robot while still keeping a high level of compliance with the environment.

In the self-tuning impedance scenario, the robot’s impedance controller achieved a Cartesian impedance profiles along the direction of the motion ($xy$ plane). The maximum controllable stiffness along the primary motion axis was set to $k_{\textrm{max}} = 1000$~N/m, while the minimum controllable stiffness was set to $k_{\textrm{min}} = 300$~N/m. The values of both $k_{\textrm{max}}$ and $k_{\textrm{min}}$ were set to align with the high and low impedance scenarios for a fair comparison.

\begin{figure}[t]
  \centering
  \includegraphics[width=0.7\linewidth]{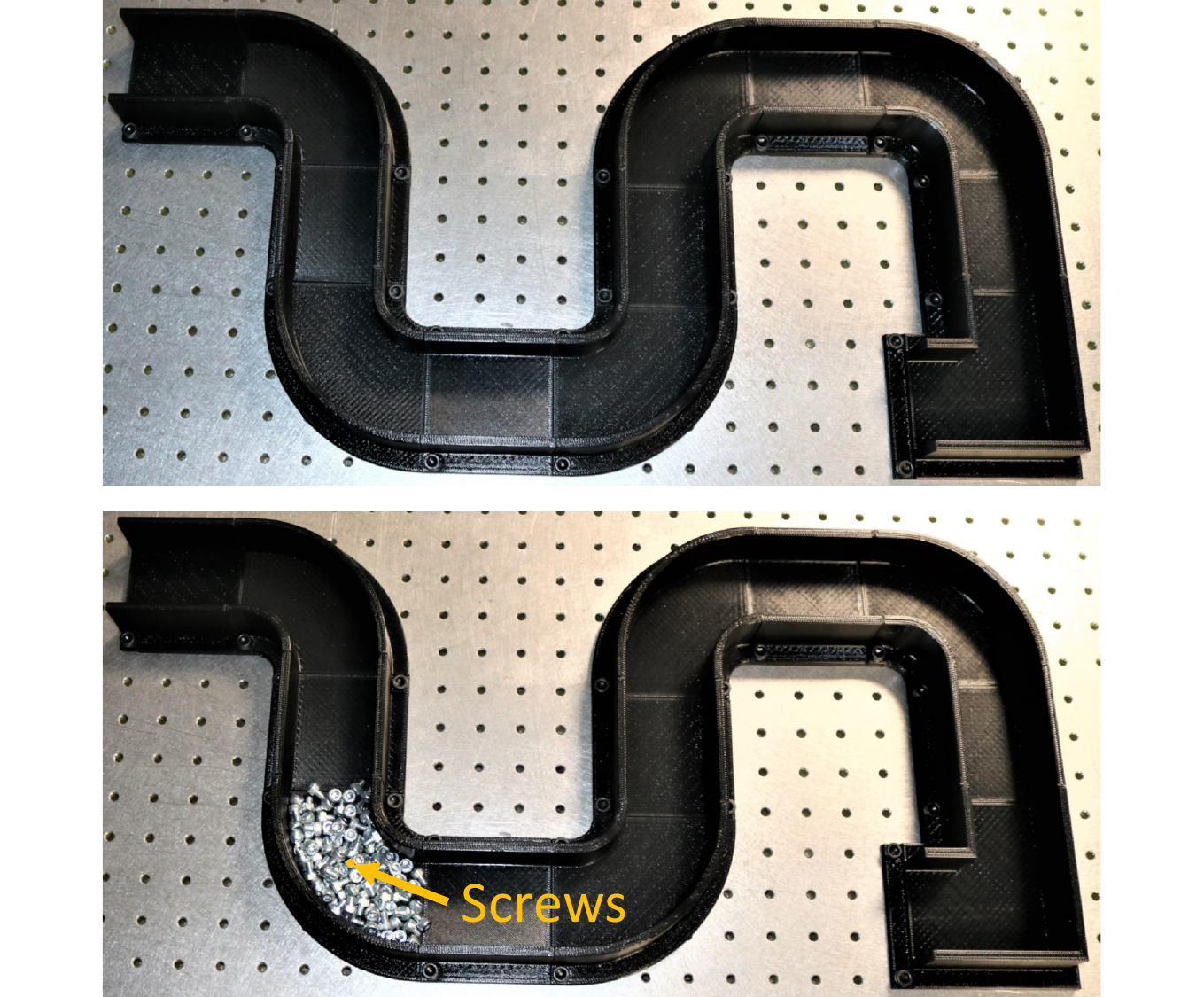}
  \caption{Experimental setup: (top) is the maze path for the first experiment. (bottom) is the maze with a cluttered condition by screws for the second experiment.}
  \label{Exp2_fig1}
  \vspace{-7mm}
\end{figure}

\begin{figure*}[ht]
  \centering
  \includegraphics[trim=0.2cm 0.0cm 0.0cm 0.0cm,clip,width=0.9\textwidth]{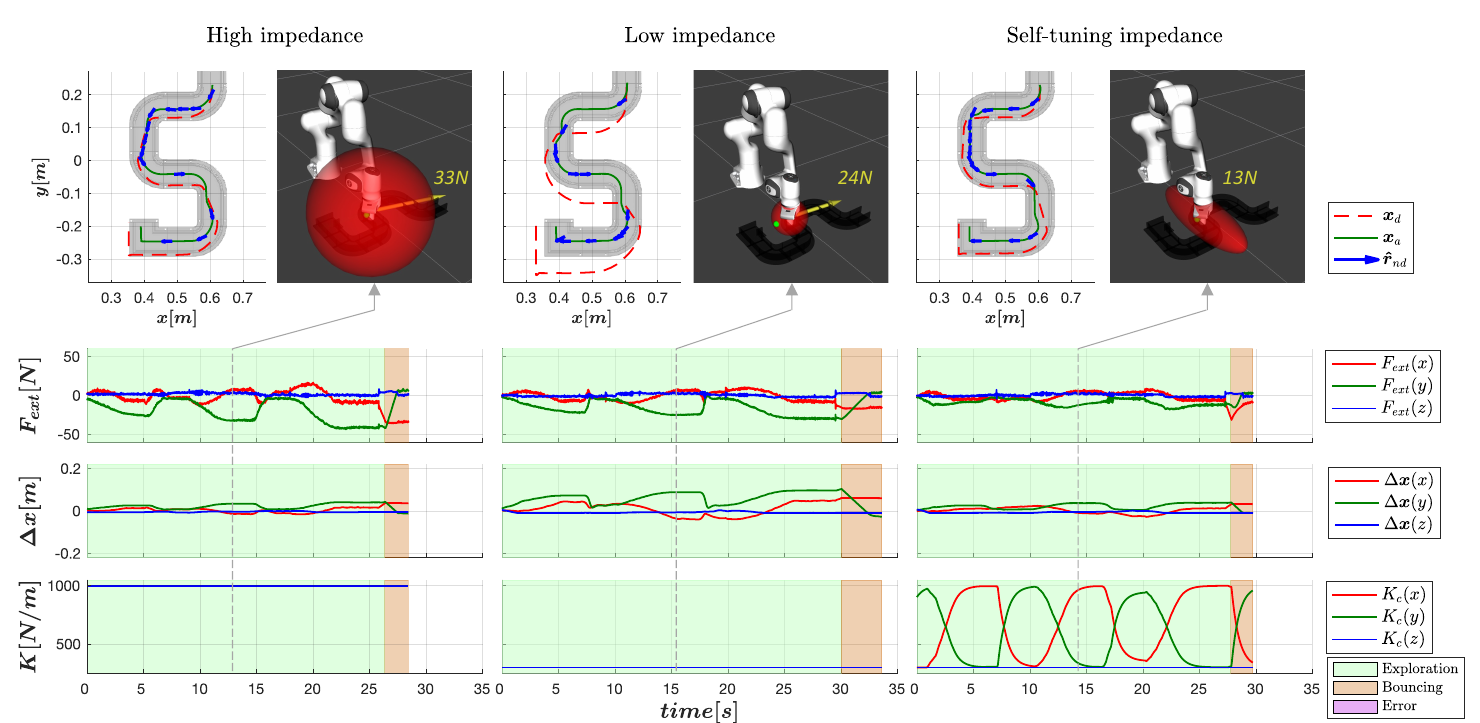}
  \caption{Comparison of three impedance profiles during exploration task: the upper row shows the maze path and the robot's trajectory. The red dash line shows the desired trajectory while the green line plotted the actual trajectory. The blue arrow shows the motion vector of the robot. Additionally, snapshots of the real robot's behavior at a similar point in the maze are provided. The red marker visualizes stiffness ellipsoid while the yellow arrow shows the interaction forces. The green dot represents the desired pose of the robot. High impedance (left) shows high interaction forces compared to the other impedance profiles. Low impedance (middle) shows larger $\Delta{\bm{x}}$. This is also visualized in the snapshots as the desired pose was far from the actual end-effector position. Self-tuning impedance (right) shows smaller $\Delta{\bm{x}}$ as in the case of high impedance and interestingly, interaction forces smaller with respect to the low impedance case. Moreover, the execution time of self-tuning impedance is comparable to high impedance while low impedance showed a slightly longer time.} 
  \label{Exp1_fig1}
  \vspace{-4mm}
\end{figure*}

We follow the FSM states sequence to describe the experiments. After the robot is sent to the starting position, the \textit{Exploration} state takes place. In the \textit{Exploration} state, an interaction with the environment was expected to happen, and therefore the self-tuning impedance unit was enabled. As the robot came into contact with the environment, the interaction planner changed the desired pose by shaping the desired velocity $v_{\textrm{constE}}$, which was set to $0.04$~m/s, when external forces exceeded the two forces thresholds set to $F_{\textrm{th\_low}} = 5$~N and $F_{\textrm{th\_high}} = 7$~N. Those forces threshold values were set with relatively high values, so as to mitigate the influence of the signal noises. Consequently, the impedance tuning allowed the robot to be rigid only in the direction of the primary motion and compliant in all the other directions, thus ensuring to make a compliant contact with the rigid environment which is the crucial information for our interaction based planner.
At the very end of the maze, an L-shaped corner was set up and the robot was trapped in a state where it was unable to move. This is because the interaction planner in the \textit{Exploration} state modifies the trajectory based on the displacement caused by the environment, and the robot was unable to detect a movable direction  colliding perpendicularly to the surface of the environment. Here, following~\eqref{eq:bouncingswitch}, with $R_{\textrm{th}} = 1$~mm and $h = 500$~ms, the FSM shifted to the \textit{Bouncing} state, since $\Delta{d} < R_{\textrm{th}}$.
In the \textit{Bouncing} state, the robot determined the possible direction of motion by scaling $v_{\textrm{constB}}$, which was set to $0.05$~m/s, based on two vectors i.e., $\bm{\hat{F}}_{\textrm{ext}}$ and $\bm{\hat{r}}_{\textrm{trend}}$. The sampling time steps to update $\bm{\hat{r}}_{\textrm{trend}}$ was set to $m = 2000$. 
Fig.~\ref{Exp1_fig1} shows the experimental result of exploration task with the three impedance profiles described above. In all the scenarios, the proposed interaction planner enabled the robot to complete the exploration task based on the contacts with the environment. 
However, it was observed from the results that the interaction behavior appears different for each impedance profile. To compare the results among the different controllers, we also define the expected completion time (CT), assuming the optimal collision-free path exploration at $\dot{x}_d$, and its exploration distance (ED). For the maze configuration illustrated in Fig.~\ref{Exp2_fig1} these two values are $26.0$\;s and $1.06$\;m, respectively.

Fig. 6 (left) shows the data obtained while maintaining high impedance profiles ($1000$~N/m). Although, this setup showed high desired pose tracking capability ($\Delta\bm{x}_{\textrm{max}}\approx0.04$~m), where $\Delta\bm{x}$ is the tracking error, the high impedance profiles resulted in higher average interaction forces $\textrm{F}_{\textrm{avg}}\approx23$~N due to the rigid contact with the environment. In terms of performances, the actual CT was equal to $31$\;s and ED $1.16$\;m.

Fig. 6 (middle) shows the plots of the trial with low impedance profiles ($300$~N/m). Despite this setup provides compliance, relatively higher $\textrm{F}_{\textrm{avg}}\approx18$~N were observed due to the larger value of $\Delta\bm{x}_{\textrm{max}}\approx0.10$~m. The possible cause of larger $\Delta\bm{x}$ is as follows. For low impedance profiles, the interaction forces gradually increased. Since the force thresholds used by the proposed interaction planner to change the trajectory was set to relatively high values, this allowed for a larger $\Delta\bm{x}$ before changing the trajectory. Another drawback given by the lower impedance profiles, was that the robot was unable to compensate for $\Delta\bm{x}$ due to its lower tracking ability for desired pose, thus highlighting a less desirable behavior. Moreover, the actual CT was $36$\;s and ED was $1.13$\;m.

Fig. 6 (right) depicts the data employing the self-tuning impedance method. $k_{\textrm{max}} = 1000$~N/m was set along the primary motion axis while $k_{\textrm{min}} = 300$~N/m was set in the other axis. Since the high impedance was maintained in the motion direction, the error was as small as the high impedance profiles. Furthermore, $\textrm{F}_{\textrm{avg}}\approx11$~N 
were lower than the low impedance profiles because of the low impedance in non primary motion direction and the smaller $\Delta\bm{x}_{\textrm{max}}$ ($\approx0.04$~m). Additionally, the actual CT was $32$\;s and ED was $1.13$\;m. CT is comparable to the stiff case, and 11\% lower with respect to the compliant one. On the other hand, ED is comparable to the compliant case and 2.5\% lower with respect to the stiff case. Taking into account the optimal values, the increase in CT is 23\% while for ED is 7\%. These considerations demonstrate the exploration efficiency of our algorithm despite our interaction planner seeks real-time haptic interaction/collision with unknown environments.

\begin{figure*}[hbt!]
  \centering
  \includegraphics[trim=0.2cm 0.0cm 0.0cm 0.0cm,clip,width=0.9\textwidth]{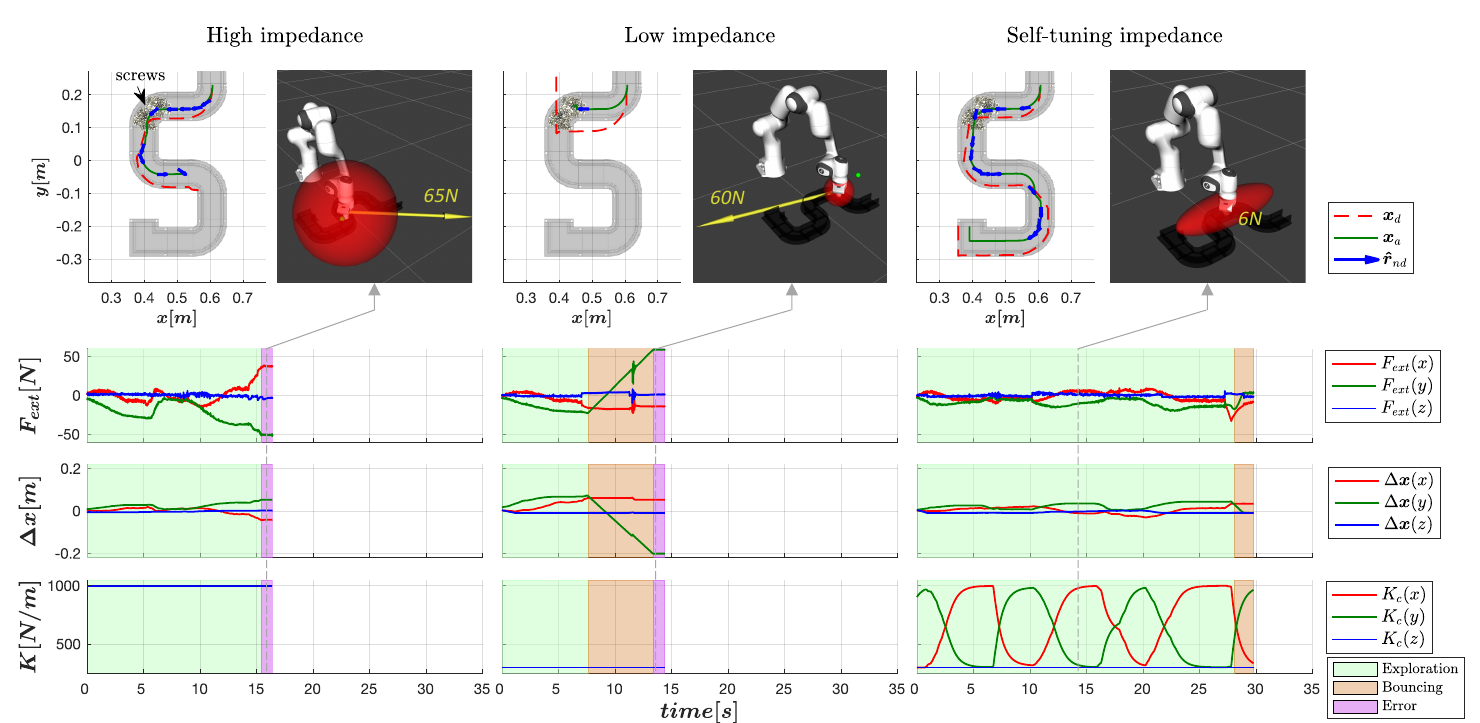}
  \caption{Comparison of three impedance profiles during exploration task in cluttered condition: as depicted in the first row, the maze path was cluttered by screws. In high impedance (left) and low impedance (middle) scenarios, robot's motion was aborted due to high interaction forces ($>60$~N). The self-tuning impedance scenario (right) showed its robustness as it completed the task execution. The interaction forces were significantly lower with self-tuning impedance compare to the other impedance profiles as visualized in the snapshots.}
  \label{Exp2_fig2}
  \vspace{-4mm}
\end{figure*}

\begin{figure*}[ht]
  \centering
  \includegraphics[trim=0.1cm 0.1cm 0.0cm 0.0cm,clip,width=\textwidth]{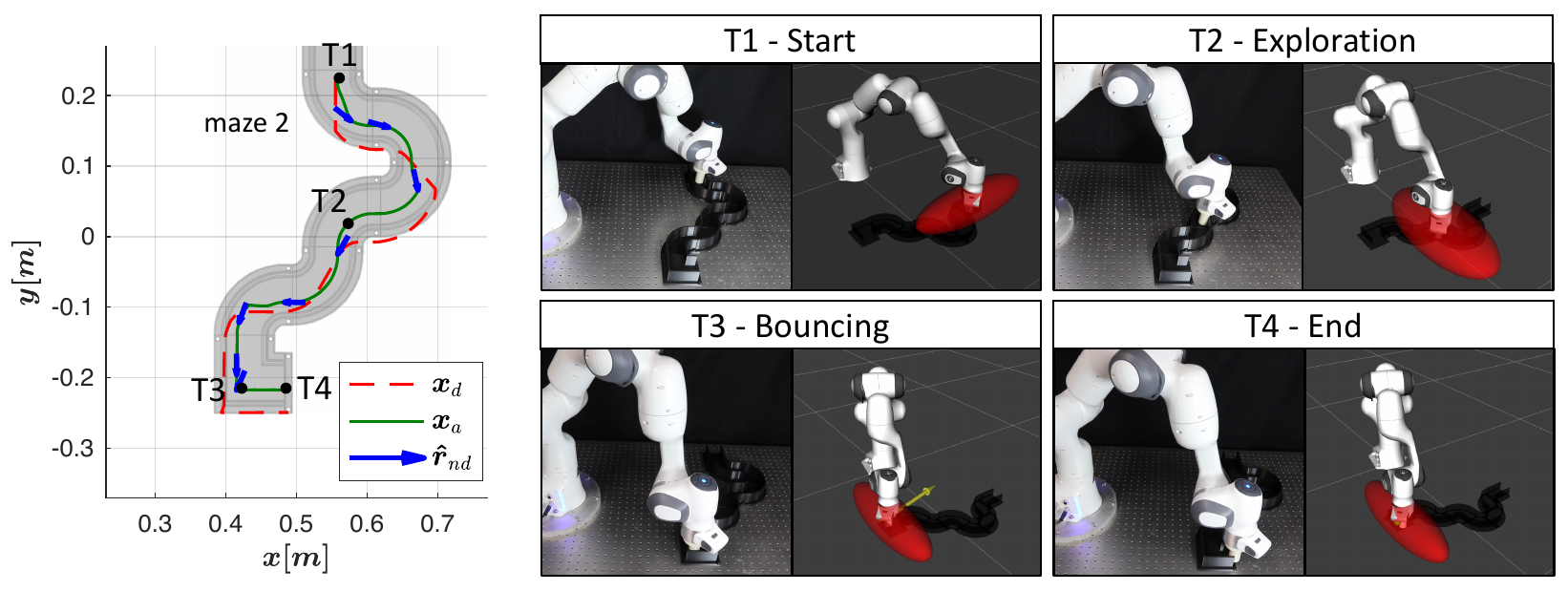}
  \caption{To show the algorithm scalability with different constrained environments, the interaction planner has been tested with mazes with different features. Here, \textit{maze 2} is constructed mainly by curved parts. As can be inferred by the $xy$ plane plot (left), the interaction planner frequently changed the motion direction $\bm{\hat{r}_{\textrm{nd}}}$, so as to adapt to the continuous changes of the maze path. On the right four different frames show the real robot and its 3D visualization enriched with impedance values (red ellipsoid) and interaction forces (yellow arrow).}
  \label{Exp3_fig1}
  \vspace{-2mm}
\end{figure*}

\begin{figure*}[ht]
  \centering
  \includegraphics[trim=0.1cm 0.1cm 0.0cm 0.0cm,clip,width=\textwidth]{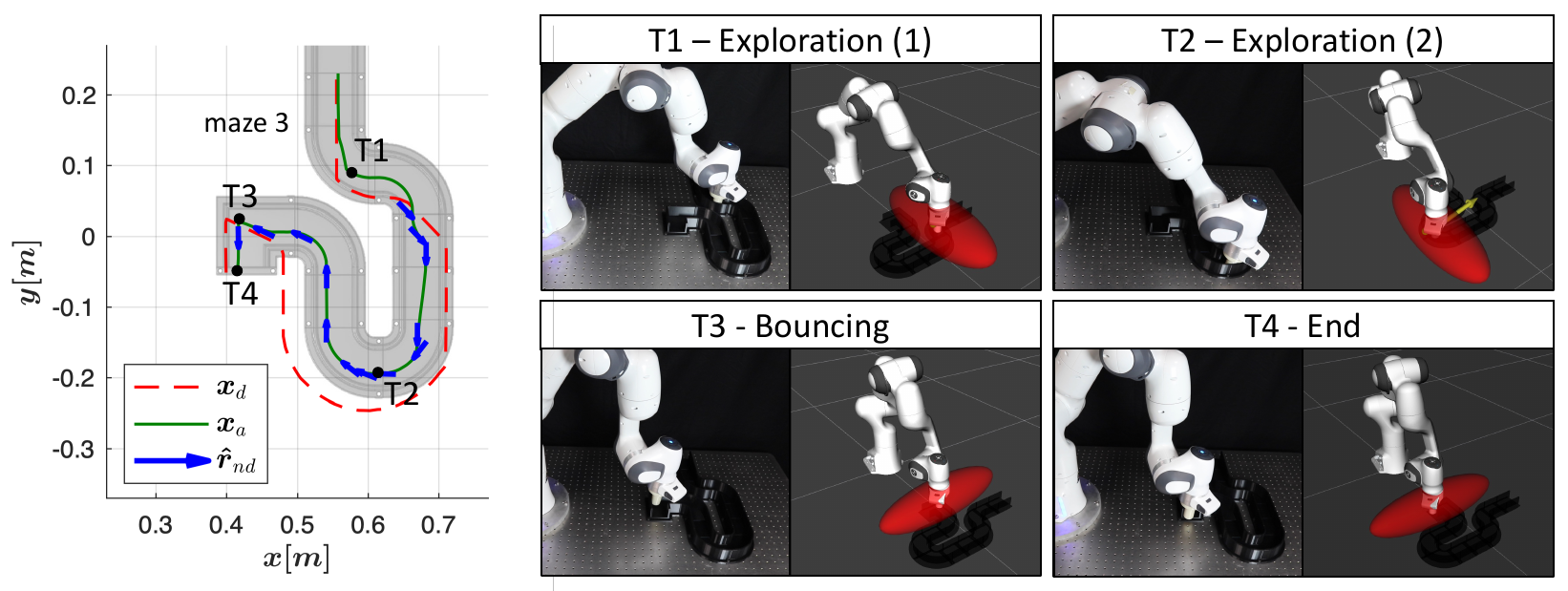}
  \caption{Similarly as Fig.~\ref{Exp3_fig1}, \textit{maze 3} shows another scenarios where the U-shape requires dynamic changes of the trajectory. Nevertheless, the presented interaction planner was able to successfully execute the exploration task.}
  \label{Exp4_fig1}
  \vspace{-4.5mm}
\end{figure*}

In the second experiment, the setup was modified by adding cluttered condition, to add soft resistance to the robot interaction. Possible applications of the proposed method, e.g., insertion tasks/pipe inspection, where some degree of resistance is expected to occur due to the cluttered conditions. It is desirable for a robot to be capable of performing the task regardless of such disturbances. As shown in Fig.~\ref{Exp2_fig1}, the robot performed an exploration task in the same maze as in the first experiment but with screws added along the path.
As illustrated in Fig. \ref{Exp2_fig2}, the experimental result showed that the task was completed only by self-tuning impedance scenario. Figs. \ref{Exp2_fig2} (left) and (middle) are the data with high and low impedance profiles, respectively. In high impedance scenario, the sum of interaction forces increased gradually and reached the maximum force threshold ($60$~N). Hence, the FSM transited to \textit{Error} state which aborted the robot's task execution. In low impedance case, due to lower tracking capability, the robot's motion was disturbed by screws as shown in larger $\Delta\bm{x}_{\textrm{max}}\approx0.07$~m. Approximately at $t = 8$~s, the interaction planner shifted to \textit{Bouncing} behavior due to small actual pose change and generated high interaction forces. Eventually, the robot's task execution was aborted because of the high interaction forces ($> 60$~N). Fig.~\ref{Exp2_fig2} (right) shows the robot's behavior in self-tuning impedance scenario. The stiffness ellipsoid design was robust to disturbances in the direction of motion, while being compliant in other directions, thus suppressing high interaction forces.

\subsection{Algorithm scalability}\label{subsec:exp_3}
In the third experiment, the robot performed an exploration task in two other mazes. The maze designs were different from the one in the first experiment in order to show the scalability of the proposed method. When a robot performs tasks in unstructured environments or pHRC scenarios, a robot's work space is expected to have a wide variety of environmental models. Therefore, it is desirable for a robot to be capable of performing the task despite various changes of the environment model as the maze was reconstructed in multiple ways. 
As shown in Figs. \ref{Exp3_fig1} and \ref{Exp4_fig1}, the results of this experiment showed that the proposed planner with self-tuning impedance method can successfully perform the exploration tasks regardless of its variance in the environmental model. In the left side of the figures, the maze paths and the robot's trajectory are plotted. It can be seen that the trajectory was adapted for each maze and the robot was following with $\Delta{\bm{x}_{\textrm{max}}}\approx0.04$~m in \textit{maze~2} and $\Delta{\bm{x}_{\textrm{max}}}\approx0.06$~m in \textit{maze~3}. The right part of the figure shows four frames of the real robot's behavior and its 3D visualization enriched with information about its desired pose planned by the interaction planner (green dot), and about the stiffness ellipsoid that was changing its shape/direction accordingly.

\section{Conclusion and discussion}\label{sec:conclusion}
This study presented a novel method to plan the robot's trajectory based on interaction with the environment and tune impedance profiles accordingly. The robot had to execute exploration tasks without prior knowledge of the environment. It performed the task within the FSM by changing its strategies according to robot's state. The proposed interaction planner mainly used \textit{Exploration} and \textit{Bouncing} strategies to perform trajectory planning. The experimental results have shown that the presented interaction planner can successfully plan trajectories in feasible directions. In addition, a compliant interaction was ensured by tuning impedance despite unexpected changes in the environment. The forces thresholds of the interaction planner should be predefined at reasonable values. This part is not yet automated and was adjusted through preliminary trials. However, it was not in the focus of this study because we considered interacting only with rigid environment, and hence, the concern was mainly signal noises.

The strength of the proposed method relies in the adaptation capabilities to the environmental changes, not relying on learning techniques \cite{li2017adaptive,yang2018dmps} in both interaction planner and self-tuning impedance. In this study, the robot was required to execute exploration tasks without prior knowledge of the environment. Additionally, several variations of environment models were prepared as in the third experiment. In such a scenario, a robot needs continuous model adaptation in the learning scheme which may not be practical due to scarce data availability. In contrast, our proposed method completed exploration tasks in several maze paths only by interacting with the environment without the need for such model adaptation. 

In addition, we were able to exploit the impedance design in the exploration tasks. The geometric ellipsoid of the self-tuning impedance method is similar to the one of human impedance design which elongates impedance only in the required directions \cite{burdet2001central}. The second experiment well demonstrated the advantage of our impedance design. In high impedance scenario, since the impedance was also high in the unnecessary axis, the robot generated large energy by interacting with rigid environments which resulted in aborting the task execution, while lower stability against disturbances was observed in low impedance scenario. In self-tuning impedance scenario, by maintaining the high impedance at the contact point, this strategy helped to stabilize the system against disturbances. The stability of the system can be analyzed by the tank-based system passivity observer as already demonstrated in other applications of variable impedance control~\cite{wu2020framework}. Furthermore, the interaction planner and self-tuning impedance unit were closely related in a sense that impedance was tuned according to the planned trajectory. In this way, a compliant interaction was ensured during collisions. This behavior, planning trajectory and impedance simultaneously, is mimicking human behavior of contact tasks \cite{ganesh2010biomimetic}. Therefore, our proposed framework has shown robustness for robotic autonomous exploration task in unknown environments. In future works, we will consider the extension of our proposed methodology to 3D mazes and explore its adaptation to bio-inspired haptic-based navigation approaches based on the behavior of vision-impaired insects and animals such as worms and moles.

\bibliographystyle{ieeetr}
\bibliography{biblio.bib}

\end{document}